\documentclass[letterpaper, 10 pt, conference]{ieeeconf}  

\IEEEoverridecommandlockouts                              

\usepackage{graphicx}      
\usepackage{algorithm} 
\usepackage{algpseudocode} 
\usepackage{varwidth} 
\usepackage{amsfonts}
\usepackage{amsmath}
\usepackage{url}

\usepackage[caption=false,font=footnotesize]{subfig}

\graphicspath{{figures/}}
\title{\LARGE \bf
Optimization Model for Planning Precision Grasps with Multi-Fingered Hands
}

\author{Yongxiang Fan, Xinghao Zhu, Masayoshi Tomizuka
\thanks{Yongxiang Fan, Xinghao Zhu, and Masayoshi Tomizuka are with Department of Mechanical Engineering, 
        University of California, Berkeley, Berkeley, CA 94720, USA
        {\tt\small {yongxiang\_fan, zhuxh, tomizuka}@berkeley.edu}}%
}

\begin{document}
\maketitle
\thispagestyle{empty}
\pagestyle{empty}

\begin{abstract}
Precision grasps with multi-fingered hands are important for precise placement and in-hand manipulation tasks. Searching precision grasps on the object represented by point cloud, is challenging due to the complex object shape, high-dimensionality, collision and undesired properties of the sensing and positioning.  
This paper proposes an optimization model to search for precision grasps with multi-fingered hands. The model takes noisy point cloud of the object as input and optimizes the grasp quality by iteratively searching for the palm pose and finger joints positions. The collision between the hand and the object is approximated and penalized by a series of least-squares. The collision approximation is able to handle the point cloud representation of the objects with complex shapes. 
The proposed optimization model is able to locate collision-free optimal precision grasps efficiently. The average computation time is 0.50 sec/grasp. The searching is robust to the incompleteness and noise of the point cloud. The effectiveness of the algorithm is demonstrated by experiments. The experimental video is available at~\cite{website}. 
\end{abstract}

\section{Introduction}
Grasping using multi-fingered robot hands have obtained increasing focus in recent years. On one side, the increase of joints offers more degree of freedoms (DOFs), which intrinsically provides more manipulability, makes the grasp more dexterous than parallel-jaw grippers. Moreover, by introducing more contact points, the grasp with multi-fingered hands is able to resist larger disturbances during the grasping and manipulation tasks. On the other side, the increase of DOFs also makes the calculation of contact points more complex.

As one category of grasps with multi-fingered hands, precision grasp has great importance in grasping the small/flat objects or executing the high-precision in-hand manipulation tasks. In these tasks, a robotic hand uses the fingertips to contact with the object and manipulate the object with the force/torque produced from these contacts. With the maximum joints in the kinematic tree, the fingertips provide enough dexterity and generate forces in different directions for object manipulation and disturbance rejection.  

The realization of precision grasps, however, is challenging due to the complex shape of the objects, high-dimensionality of the planning problem, collision avoidance requirements during grasp searching and execution, and the robustness to various uncertainties. 

To plan optimal grasps, pure mathematical model, as well as learning based methods, could be applied. Using methods from the prior category, a grasp quality metric is always helpful. The idea was proposed in~\cite{ferrari1992planning} as the largest-minimum resisted wrench of the grasp wrench space. The metric describes the maximum disturbance that the grasp can resist. Nevertheless, the computation is excessively heavy to search multiple contacts across the object, which limits the application to real-time applications. 
To address this issue,~\cite{hang2016hierarchical} formulated the grasp planning problem using the hierarchical fingertip space (HFTS) and optimized the largest-minimum resisted wrench~\cite{ferrari1992planning} with stochastic hill climbing algorithm. The optimization requires offline object processing, and the collision was addressed by an optimize-then-prune method. 
To accelerate the computation, a part-based grasp planning method based on the Reeb graph of the object was proposed in~\cite{aleotti2011part}. However, the Reeb graph is slow to build, and it assumes only one part will be grasped for on object, whereas the optimal grasp can expand on multiple parts, especially for small objects. 
In~\cite{vahrenkamp2018planning}, the grasps were sampled from the mean curvature skeleton of the object with the manually-designed heuristics. The skeleton is computationally heavy, and the collision was addressed by simply moving backward against the approaching direction. 

Learning-based methods have been proposed to address the efficiency issue. An example based planning framework was proposed in~\cite{dang2012semantic} to generate stable grasps for specific object manipulation tasks. The example database, however, requires extensive human demonstration, and the learning is valid only for those objects similar to the database instead of the arbitrary shape. 
Similarly, a grasp strategy was learned in~\cite{huang2013learning} from the previous grasps on the same object. The learned policy cannot adapt to unknown objects. 
To increase the generality of the database to unknown objects, a hierarchical reinforcement learning approach was introduced in~\cite{osa2016experiments} for appropriate grasp type/location selection. The grasps were generalized to unknown objects by extracting the patches around the grasps and mapped the patch as well as the learned grasp to the unknown object by Iterative Closest Point (ICP). Despite the generalization to the unknown object, the searching can be excessively heavy for a large database. Moreover, the database only contains local geometric information related to the grasp, while a successful grasp may also rely on the global geometry for quality computation and collision avoidance.
Supervised learning based grasp was proposed in~\cite{varley2015generating} to learn grasps for multi-fingered hands. Due to the high dimensionality and the shape diversity, the learning module was combined with eigen-grasp planner~\cite{ciocarlie2007dexterous}. It took 16 secs to generate a grasp. 

To address the learning complexity with multi-fingered hands, we proposed a finger splitting algorithm for grasp generation based on the grasps from parallel grippers~\cite{fan2018real}. The algorithm splits fingers by alternatively optimizing the palm pose and the contact positions. Instead of searching for new contacts to avoid the collision or reduce misalignment, the algorithm terminated if there was collision or misalignment between contact normal and fingertip normal. Moreover, the palm has less freedom to adjust with fixed contacts if the hand is under-actuated or have low DOFs, thus the searching range is limited. 

In this paper, we propose an optimization model to search for optimal grasps while avoiding collision with multi-fingered hands. The optimization model formulates the planning problem into a gradient-based optimization. The optimization is solved by iterating between a palm pose optimization (PPO) and a joint position optimization (JPO). PPO searches for desired palm pose to optimize the geometrically related object quality metrics with fixed finger joints. In contrast, JPO searches for desired finger joint positions to optimize both the object quality and hand manipulability with fixed palm pose. 

The contributions of this paper are as follows. First, by formulating the problem as a gradient-based optimization with a random start. Moreover, the collision is actively penalized and avoided within the optimization instead of pruning the collided grasps after the optimization. With the customized penalization, the iterative PPO-JPO changes the non-convex planning problem into a sequence of least-squares, and the average computation time is 0.5 sec/grasp. Third, the algorithm is robust to sensing uncertainties and noises by employing contact patches rather than individual contact points. The effectiveness of the algorithm is verified by both the simulation and the experiment. The experimental videos are available at~\cite{website}.

The remainder of this paper is as follows. The formulation of the planning problem is stated in Section~\ref{sec:problem_statement}. Section~\ref{sec:framework} and Section~\ref{sec:ppojpo}  present the optimization model and its solution by iterative PPO-JPO. Simulation and experiment are introduced in Section~\ref{sec:sim_exp}. Section~\ref{sec:conclusion} concludes the paper and introduces the future work.

\section{Problem Statement}
\label{sec:problem_statement}
\begin{figure}[tb]
	\begin{center}
		\includegraphics[width=1.2in]{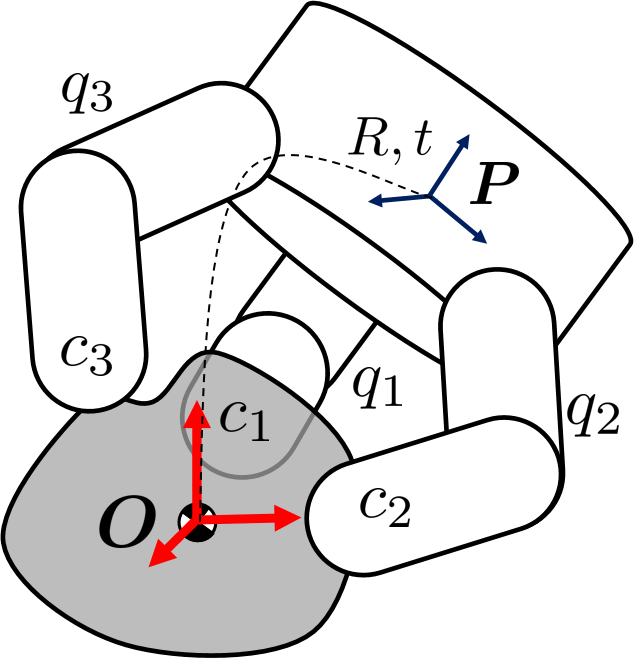}
		\caption{Illustration of grasp planning problem using a three-fingered hand. }
		\label{fig:object_hand}
	\end{center}
\end{figure}
We consider the planning of precision grasps with multi-fingered hands. In the precision grasp mode, the hand contacts with the object by fingertips to gain most dexterity and increase the grasping manipulability. Precision grasps are necessary if the object to be grasped is flat or require further precision operations. 
An example of precision grasp with a three-fingered hand is shown in Fig.~\ref{fig:object_hand}. The contacts that connect the hand $\mathcal{F}$ and the object $\partial \mathcal{O}$ are denoted as $\boldsymbol{c} = \left[c_1,..., c_{N_c}\right]$, where $N_c$ is the number of contacts. Different contacts are associated by the hand configuration characterized by the palm pose $(R\in SO(3),\boldsymbol{t}\in \mathbb{R}^3)$ and the joint angles $\boldsymbol{q} = \left[q_1,...,q_{N_c}\right]$, where $q_i\in \mathbb{R}^{N_{jnt,i}}$ is joint angles of the $i$-th finger and  $N_{jnt,i}$ is the number of joints for this finger. 

Given the object $\partial \mathcal{O}$, the grasp planning is to determine the grasp $\mathcal{G} = \{\boldsymbol{c}, (R,\boldsymbol{t}), \boldsymbol{q}\}$ to successfully lift the object. To be more specific, 
\begin{subequations}
	\label{eq:general_form}
	\begin{align}
	\max_{R, \boldsymbol{t}, \boldsymbol{q},\boldsymbol{c}} &\  Q(\boldsymbol{c}, \boldsymbol{q}) \label{eq:general_cost}\\
	s.t. \quad 
	& \boldsymbol{c} \in FK(\mathcal{F}_t; R,\boldsymbol{t}, \boldsymbol{q}) \label{eq:general_FK}\\
	& \boldsymbol{c} \in \partial \mathcal{O} \label{eq:general_surface} \\ 
	& dist(FK(\mathcal{F};R,\boldsymbol{t}, \boldsymbol{q}), \partial \mathcal{O}) \geq 0 \label{eq:collision}\\
	& q_i \in [q_{\text{min},i}, q_{\text{max},i}] \label{eq:general_limit} \quad i = 1\cdots N_c
	\end{align}
\end{subequations}
where $Q(\boldsymbol{c},\boldsymbol{q})$ denotes the grasp quality including the $Q_{center}(\boldsymbol{c},\partial\mathcal{O})$, $Q_{jcenter}(\boldsymbol{q})$ and $Q_{align}(n_c, n_f)$.  
$Q_{center}$ is the distance between the centroid of the contact polygon and the object’s center of mass. With minimal distance, the moment of gravitational force is reduced due to the small moment arm. $Q_{jcenter}$ is the derivation from the center of the joints. $Q_{align}$ is the misalignment between the contact normal $n_c$ and fingertip normal $n_f$. A large misalignment error implies the exerted force may be outside of the friction cone.

Constraint~(\ref{eq:general_FK}) is the kinematic constraint that connects the hand configuration with the contacts, where $\mathcal{F}_t$ represents the fingertip surfaces and $FK(\mathcal{F}_t;  R,\boldsymbol{t}, \boldsymbol{q})$ denotes the forward kinematics parameterized by hand configuration. Constraint~(\ref{eq:general_surface}) restricts the contacts to object surface $\partial O$, (\ref{eq:collision}) denotes that the hand surface $\mathcal{F}$ parameterized by $(R,\boldsymbol{t}, \boldsymbol{q})$ should not collide with the object, 
and (\ref{eq:general_limit}) shows the joint limits. 
With the palm pose, joints and contacts as optimization variables, forward kinematics~(\ref{eq:general_FK}), complex object surface~(\ref{eq:general_surface}) and collision~(\ref{eq:collision}) as constraints,  Problem~(\ref{eq:general_form}) becomes a high-dimensional nonlinear programming. 

This paper proposes an optimization model to solve Optimization~(\ref{eq:general_form}). The details of the algorithm are introduced below. 

\section{Framework}
\label{sec:framework}
\begin{figure}[tb]
	\begin{center}
		\includegraphics[width=3.3in]{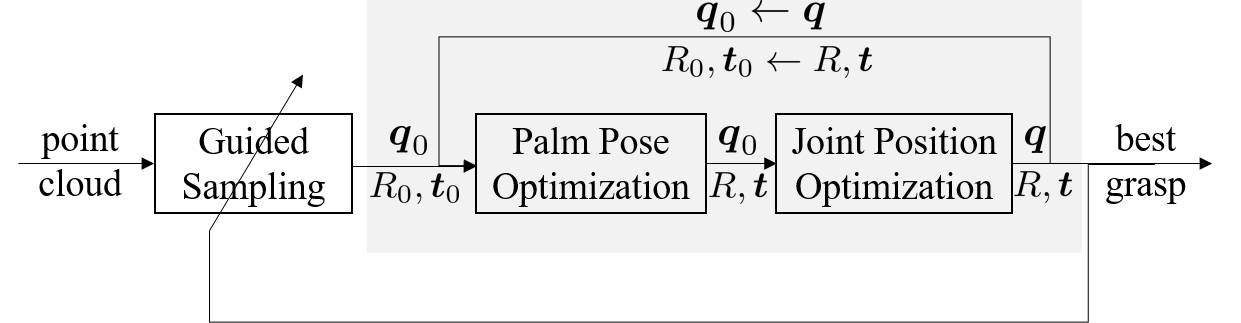}
		\caption{Framework of the grasp planning algorithm. The inner loop iterates between the palm pose optimization and the joint position optimization, as shown in gray. The outer loop initialize the hand configuration and restarts the inner loop.}
		\label{fig:framework}
	\end{center}
\end{figure}
The framework of the proposed grasp planning algorithm is shown in Fig.~\ref{fig:framework}. It contains two loops. the inner loop iterates the palm pose optimization (PPO) and the joint position optimization (JPO) and is called the iterative PPO-JPO. The outer loop samples hand configuration and restarts the iterative PPO-JPO. 
The PPO algorithm fixes joints, $\boldsymbol{q} = \boldsymbol{q}_0$, and searches for the palm pose $(R,\boldsymbol{t})$ by optimizing the grasp quality $Q_{com}(\boldsymbol{c}(R,\boldsymbol{t},\boldsymbol{q}_0),\partial \mathcal{O}) + Q_{algin}(n_c,n_f)$ while minimizing the collision with the object and the ground. The JPO algorithm fixes the palm pose $(R,\boldsymbol{t}) =(R_0,\boldsymbol{t}_0)$ while searches for the joints $\boldsymbol{q}$ by optimizing the $Q_{com}(\boldsymbol{c}(R_0, \boldsymbol{t}_0, \boldsymbol{q}),\partial \mathcal{O})+Q_{jc}(\boldsymbol{q}) + Q_{align}(n_c, n_f)$ and minimizing the collision. The iterative PPO-JPO finds a local optimum of the optimization~(\ref{eq:general_form}). 

The guided sampling in the outer loop is introduced from~\cite{fan2018grasp} to avoid the iterative PPO-JPO being trapped in poor-performed local optima. It employs the K-means clustering of the object surface and places the hand onto the cluster centers. The success rate of locating a high quality collision-free grasp is different due to the varying local geometries. The guided sampling ranks different clustering centers based on the success rate and the center with higher success rate will be sampled more often. 
The details of the guided sampling are neglected in this paper for simplicity.

\section{Iterative PPO-JPO}
\label{sec:ppojpo}
\subsection{Constraint Relaxation}
Optimization~(\ref{eq:general_form}) is an abstract formulation to demonstrate the considerations in grasp planning. We relax some of the constraints to locate the grasps smoothly in the space occupied by the object, including the surface constraint~(\ref{eq:general_surface}) and the collision constraint~(\ref{eq:collision}). More specifically, 
\begin{subequations}
	\label{eq:relaxed_form}
	\begin{align}
	\max_{R, \boldsymbol{t}, \boldsymbol{q},\boldsymbol{c}} &\  Q(\boldsymbol{c}, \boldsymbol{q}) - w(E_{col} + E_{cls}) \label{eq:relaxed_cost}\\
	s.t. \quad 
	& \boldsymbol{c} = NN_{\partial \mathcal{O}}(FK(\mathcal{F}_t; R,\boldsymbol{t}, \boldsymbol{q}))\label{eq:relaxed_contacts}\\
	& q_i \in [q_{\text{min},i}, q_{\text{max},i}] \label{eq:relaxed_limit} \quad i = 1\cdots N_{f}
	\end{align}
\end{subequations}
where $NN(\bullet)_{\partial \mathcal{O}}$ in~(\ref{eq:relaxed_contacts}) denotes the nearest neighbor search of $\bullet$ on object surface. $E_{col}(R,\boldsymbol{t}, \boldsymbol{q})$ corresponds to (\ref{eq:collision}) and penalizes the collision violation. $E_{cls}(\boldsymbol{c}, FK(\mathcal{F}_t; R,\boldsymbol{t}, \boldsymbol{q}))$ and the constraint~(\ref{eq:relaxed_contacts}) together correspond to (\ref{eq:general_FK},\ref{eq:general_surface}) and penalize the distance between the contact and the fingertip, and $w$ is the penalty weight for constraint violation. 

Optimization~(\ref{eq:relaxed_form}) is a relaxed formulation of~(\ref{eq:general_form}) to reduce the nonlinearities introduced by~(\ref{eq:general_FK}) and~(\ref{eq:collision}). Despite the relaxation, the direct optimization of $R\in SO(3), \boldsymbol{t},\boldsymbol{q}$ is challenging in the following aspects: 1) the palm orientation is constrained in special orthogonal group $SO(3)$, 2) $R, \boldsymbol{t},\boldsymbol{q}$ are searched based on the quality determined by $FK(R,\boldsymbol{t},\boldsymbol{q})$ and contacts in complex $\partial \mathcal{O}$. 

\subsection{Incremental Search and Point Representation}
To enable the gradient-based search on the object space, we instead search incrementally on $\delta R, \delta \boldsymbol{t}, \delta \boldsymbol{q}$, where $\delta R, \delta \boldsymbol{t}$ denote the \textit{transformation} of palm and $\delta \boldsymbol{q}$ denotes the joint displacement. The problem is further simplified by sampling the \textit{current} hand surface  $\mathcal{F}$ into points $\{p_k, n_k^p\}_{k=1}^{N_p}$, where $p_k\in\mathbb{R}^3,n_k^p\in\mathbb{S}^2$ denote the $k$-th points and normals on hand surface pointing outwards. Similarly, the object surface $\partial\mathcal{O}$ is sampled into points $\{o_k, n_k^o\}_{k=1}^{N_o}$. We retrieve all those $\{p_k\}_{p_k\in\mathcal{F}_t}$ within $\mathcal{F}_t$ and then search the nearest neighbor on the object surface $NN_{\partial \mathcal{O}}(\{p_k\}_{p_k\in\mathcal{F}_t})$ to find corresponding points $\{o_k\}_{o_k\in\mathcal{I}}$. 
The mean values of $\{p_k\}_{p_k\in\mathcal{F}_t}$ and $\{o_k\}_{o_k\in\mathcal{I}}$ are denotes as  $\boldsymbol{p}_{f}=\left[p_{f_1},...,p_{f_{N_c}}\right]$ and $\boldsymbol{c} =\left[c_{1},...,c_{{N_c}}\right]$. 

With the point representation, $Q_{com}(\boldsymbol{c},\partial\mathcal{O})$ becomes:
$$Q_{com}(\boldsymbol{c},\partial\mathcal{O}) = -\sum_{i=1}^{N_c}((\bar{p}_{f_i} - p_{com})^Tn_{\perp})^2,$$ 
where $\bar{p}_{f_i} = \delta Rp_{f_i} + \delta \boldsymbol{t} + \delta RJ_{f_i}^v(q_i)\delta q_i$ is the fingertip position for finger $i$ after transformation, $J_{f_i}^v\in \mathbb{R}^{3\times N_{jnt,i}}$ is the translational Jacobian matrix at $q_i$ with $N_{jnt,i}$ denoting the number of joints in the $i$-th finger, $p_{com}$ is the object center point, and $n_{\perp}$ is the normal vector of the polygon formed by fingertips. Fingertip positions are used to replace the contacts to avoid searching on object surface. This replacement is reasonable under the assumption that $\boldsymbol{c}\approx \boldsymbol{p}_f$. 

The quality $Q_{jc}(\boldsymbol{q})$ becomes: 
$$Q_{jc}(\boldsymbol{q}) = -\sum_{i=1}^{N_c}\sum_{j=1}^{N_{jnt,i}}(\alpha_i^j\frac{q_i^j - \bar{q}_i^j}{q_{\max,i}^j - q_{\min,i}^j})^2,$$ 
where $\bar{q}_i^j,q_{max,i}^j, q_{min,i}^j$ are the mean and limit values of the $j$-th joint in the $i$-th finger.  $\alpha_i^j$ is the weights for the $j$-th joint of the $i$-th finger. 

Quality $Q_{align}(n_c, n_f)$ becomes: 
$$Q_{align}(n_c, n_f) = -\beta^2\sum_{i=1}^{N_c}(n_{c_i}\cdot \delta Re^{(J_i^w\delta q_i)\string^} n_{f_i} + 1)^2,$$ 
where $n_{c_i},n_{f_i}$ are the normals of the $i$-th contact and fingertips, respectively, and $J_i^w\in \mathbb{R}^{3\times N_{jnt,i}}$ is the rotational Jacobian matrix for finger $i$. $\beta$ is the scaling factor of $Q_{align}$.

\begin{figure}[tb]
	\begin{center}
		\includegraphics[width=3.3in]{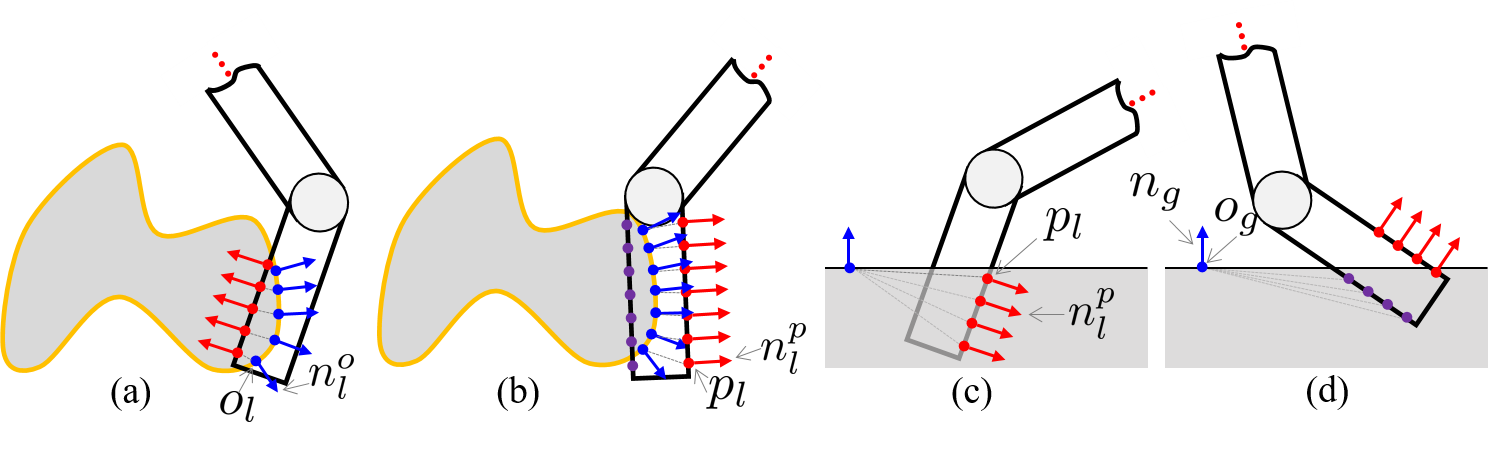}
		\caption{Illustration of collision detection and formulation. (ab) Hand-object collision. Collided points are obtained by transforming object points to bounding boxes of fingers and check the inclusion. Collision types are determined by the sign of $\sum n_l^p\cdot o_l$. (a) shows the inner side contact and (b) shows the outer side contact. For the outer side contact, $p_l$ is replaced by purple points. (cd) Hand-ground collision. }
		\label{fig:collision}
	\end{center}
\end{figure}

The formulation of collision penalty $E_{col}$ is approximated in this paper to accelerate the computation (Fig.~\ref{fig:collision}), though a rigorous formulation can be found in~\cite{schulman2013finding}. 
With the approximation shown in ~Fig.~\ref{fig:collision}, collision penalty $E_{col}$ becomes:  
$$E_{col} = \sum_{l=1}^{N_{col}}\|\bar{p}_l - o_l\|^2 + \sum_{l=1}^{N_{col,g}}((\bar{p}_l - o_g)^Tn_g)^2,$$ 
where $N_{col},N_{col,g}$ are numbers of collided points with the object and ground, respectively. $\bar{p}_l =\delta R p_l +\delta \boldsymbol{t} + \delta R \mathcal{J}_l(\boldsymbol{q})\delta\boldsymbol{q}$, where $\mathcal{J}_l\in\mathbb{R}^{3\times N_{jnt}}$ denoting the hand Jacobian matrix, $o_g\in\mathbb{R}^3,n_g\in\mathbb{S}^2$ are sampled points and normal of the ground. 

The penalty $E_{cls}$ is $\sum_{i=1}^{N_c}(\bar{p}_{f_i} - c_i)^Tn_{c_i}$. 

Problem~(\ref{eq:relaxed_form}) with the incremental search and point representation becomes: 
\begin{subequations}
	\label{eq:ispr_form}
	\begin{align}
	\min_{\delta R, \delta\boldsymbol{t}, \delta\boldsymbol{q}} &\  -Q_{com} - Q_{jc} - Q_{align} + w(E_{col} + E_{cls}) \label{eq:ispr_cost}\\
	s.t. \quad 
	& q_i \in [q_{\min,i}, q_{\max,i}] \label{eq:ispr_limit} \quad i = 1\cdots N_{f}
	\end{align}
\end{subequations}

Problem~(\ref{eq:ispr_form}) remains a nonlinear programming due to the coupling between $\delta R$ and $\boldsymbol{q}$. We solve it with the iterative PPO-JPO algorithm. The details of PPO and JPO are described below.

\subsection{Palm Pose Optimization (PPO)}
The PPO algorithm optimizes for $\delta R, \delta \boldsymbol{t}$ by fixing the finger joints (i.e. $\delta \boldsymbol{q} = 0$). To search on the trivial Euclidean space, the hand rotation $\delta R$ is parameterized by the axis-angle representation and approximated by $\delta R \approx I_{3\times3} + \hat{r}$, where $\hat{\bullet}$ is the matrix representation of cross product and $r\in \mathbb{R}^3$ is the angle-axis vector.  

With the fixed joint and approximation of rotation,~(\ref{eq:ispr_form}) becomes a least-square problem: 
\begin{equation}
    \label{eq:ppo}
    \min_x\| Ax - b\|_2^2 
\end{equation}
where $x = [r^T, \delta \boldsymbol{t}^T]^T \in \mathbb{R}^6$. $A =[a_{com,i}^T...a_{align,i}^T...a_{col,l}^T...a_{cls,i}^T]^T\in\mathbb{R}^{(3N_c+3|\mathcal{L}_o|+|\mathcal{L}_g|)\times6}$, with $\mathcal{L}_o$ denoting the indexes of hand-object collision and $\mathcal{L}_g$ denoting the indexes of hand-ground collision. 
$a_{com,i} = [(p_{f_i}\times n_{\perp})^T, n_{\perp}^T]$, $a_{align,i} = \beta[(n_{f_i}\times n_{c_i})^T, 0_3^T]$, $a_{col,l}$ includes hand-object collision $a_{obj,l} = w[-\hat{p}_l, I_3]$ and hand-ground collision $a_{gnd,l} = w[(p_l\times n_g)^T, (n_g)^T]$, and $a_{cls,i} = w[(p_{f_i}\times n_{c_i})^T, n_{c_i}^T]$. 

Similarly, $b = [b_{com,i}...b_{align,i}...b_{col,l}...b_{cls,i}]^T$, with 
$b_{com,i} = n_{\perp}^T(p_{com}-p_{f_i})$, $b_{align,i} = -\beta(n_{f_i}^Tn_{c_i}+1)$, $b_{col,l}$ includes $b_{obj,l} = w(p_l - o_l)$ and $b_{gnd,l}=w(p_l - o_g)^Tn_g$, and $b_{cls,i} = w(c_i - p_{f_i})^Tn_{c_i}$. 

The PPO~(\ref{eq:ppo}) is a least-squares problem and can be solved analytically by
$$x^* = (A^TA)^{-1}A^Tb.$$
The optimal palm transformation is $\delta R = e^{(x^*_{1:3})\string^}, \delta \boldsymbol{t} = x^*_{4:6}$. We update the hand configuration by $(R,\boldsymbol{t})\leftarrow (\delta R,\delta\boldsymbol{t})*(R,\boldsymbol{t})$ and start the JPO.  

\subsection{Joint Position Optimization (JPO)}
The JPO algorithm optimizes for $\delta \boldsymbol{q}$ by fixing the palm pose. 
With the fixed palm pose, JPO becomes a least-squares with constraint problem: 
\begin{subequations}
    \label{eq:jpo}
    \begin{align}
    \min_{\delta \boldsymbol{q}}&\ \| C\delta \boldsymbol{q} - d\|_2^2 \\
    s.t. \quad 
	& \delta \boldsymbol{q} + \boldsymbol{q} \in [\boldsymbol{q}_\text{min}, \boldsymbol{q}_\text{max}], 
	\end{align}
\end{subequations}
where $C = [C_{com}^T...C_{jc}^T...C_{align}^T...c_{col,l}^T...C_{cls}^T]^T\in \mathbb{R}^{(3N_c + N_{jnt}+3|\mathcal{L}_o|+|\mathcal{L}_g|)\times N_{jnt}}$
with $C_{com}=\texttt{diag}(n_\perp^TJ_i^v)$, $C_{jc} = \texttt{diag}(\alpha_i^j/(q_{\max,i}^j-q_{\min,i}^j))$, $C_{align} = \texttt{diag}(n_{c_i}^Tn_{f_i}\string^J_i^w)$, 
$c_{col,l}$ includes hand-object collision $c_{obj,l}=w\mathcal{J}_l(\boldsymbol{q})$ and hand-ground collision $c_{gnd,l}=(n_g)^T\mathcal{J}_l(\boldsymbol{q})$, and $C_{cls} = w\texttt{diag}(n_{c_i}^TJ_i^v)$. 
Similarly, $d=[d_{com,i}...d_{jc,i}...d_{align,i}...d_{col,l}...d_{cls,i}]^T\in\mathbb{R}^{3N_c+N_{jnt}+3|\mathcal{L}_o|+|\mathcal{L}_g|}$, with $d_{com,i} = (p_{com}-p_{f_i})^Tn_\perp$, $d_{jc,i}=\alpha_i^j(\bar{q}_i^j-q_i^j)/(q_{\max,i}^j-q_{\min,i}^j)$, $d_{align} = -(n_{c_i}^Tn_{f_i}+1)$, $d_{col,l}$ includes $d_{obj,l}=w(p_l - o_l)$ and $d_{gnd,l} = w(p_l - o_g)^Tn_g$, and $d_{cls,i} = (c_i-p_{f_i})^Tn_{c_i}$. 

Problem~(\ref{eq:jpo}) is a least-squares with box constraints and can be solved by either a solver or by initializing $\delta\boldsymbol{q}_0 = (C^TC)^{-1}C^Td$ and iterating between
\begin{subequations}
	\label{eq:finger_optimization_solution}
	\begin{align}
	&\delta\boldsymbol{q}_{\bar{m}} = \delta\boldsymbol{q}_m - \gamma C^T(C\delta\boldsymbol{q}_m - d)\label{eq:fo_gradient_decent}\\
	&\delta\boldsymbol{q}_{m+1} = \max(\min(\delta\boldsymbol{q}_{\bar{m}}, \boldsymbol{q}_\text{max} - \boldsymbol{q}), \boldsymbol{q}_\text{min}- \boldsymbol{q})\label{eq:fo_box_constraint}
	\end{align}
\end{subequations} 

\subsection{Iterative PPO-JPO Summary}
\begin{algorithm}[t]
	\caption{Iterative PPO-JPO Algorithm}\label{alg:isf}
	\begin{algorithmic}[1]
		\State \textbf{Input:} Initial $R_s,\boldsymbol{t}_s, \delta\boldsymbol{q}_s$, $\partial \mathcal{O}$, $\mathcal{F}$, $T_{\max}$ \label{isf:input}
		\State \textbf{Init:} $(R,\boldsymbol{t},\boldsymbol{q})\leftarrow (R_s,\boldsymbol{t}_s, \delta\boldsymbol{q}_s)$ \label{isf:init}
		\For {$t = 0, \cdots, T_{\max}$} \label{isf:paraymid}
		\State $(\boldsymbol{c},\boldsymbol{p}_f)\leftarrow \texttt{update}(FK(\mathcal{F},R,\boldsymbol{t},\boldsymbol{q}),\partial\mathcal{O})$\label{isf:update1}
		\State $\delta R^*, \delta \boldsymbol{t}^*\leftarrow \texttt{PPO}(\boldsymbol{c},\boldsymbol{p}_f)$ \label{isf:ppo}
		\State $(R,\boldsymbol{t})\leftarrow(\delta R^*,\delta \boldsymbol{t}^*)*(R,\boldsymbol{t})$\label{isf:update_rt}
		\State $(\boldsymbol{c},\boldsymbol{p}_f)\leftarrow \texttt{update}(FK(\mathcal{F},R,\boldsymbol{t},\boldsymbol{q}),\partial\mathcal{O})$\label{isf:update2}
		\State $\delta\boldsymbol{q}^* \leftarrow \texttt{JPO}(\boldsymbol{c},\boldsymbol{p}_f)$ \label{isf:jpo}
		\State $\boldsymbol{q}\leftarrow\boldsymbol{q} + \delta \boldsymbol{q}^*$ \label{isf:update_q}
		\EndFor \label{isf:paraymid2} 
		\State \Return $\{ R,\boldsymbol{t}, \boldsymbol{q}\}$
	\end{algorithmic}
\end{algorithm}

The Iterative PPO-JPO algorithm is summarized in Alg.~(\ref{alg:isf}). The algorithm is fed with the sampled hand configuration from guided sampling and the hand/object geometries~(Line~\ref{isf:input}). In each iteration, we first search the contacts by the forward kinematics and nearest neighbor (Line~\ref{isf:update1}), and run the PPO algorithm to optimize for hand pose~(Line~\ref{isf:ppo}-\ref{isf:update_rt}). The contacts are refreshed accordingly and fed into the JPO algorithm for optimized $\delta q^*$ (Line~\ref{isf:update2}-\ref{isf:jpo}). The iteration terminates after $T_{\max}$ iterations. 

\section{Simulation and Experiment}
\label{sec:sim_exp}
This section shows the simulation and experiment results. The simulation ran on a desktop with 32GB RAM and 4.0GHz CPU. The computation was conducted in Matlab and visualized in VREP. For the experiment, we used a BarrettHand BH8-282 multi-fingered hand attached to a FANUC LRMate 200iD/7L industrial manipulator for grasping. Two Ensenso N35 cameras were used to capture the point cloud of the scene.

\subsection{Parameter Lists}
The hand surface was discretized into $1798$ points and each fingertip had $216$ points sampled. The scaling factor $\alpha_{1:2}^{1:3} = 1$ and $\alpha_3^{1:2} = \sqrt{2}$ in $Q_{jc}$ to balance the gradients in different sides of the hand. The scaling factor $\beta = 0.03$ in $Q_{align}$. The collision penalty $w$ was initialized as $1.0$ and increased exponentially with factor $1.1$. The maximum iteration $T_{\max}=40$.

\subsection{Simulation Results}
\begin{figure}[t]
	\begin{center}
		\includegraphics[width=3.3in]{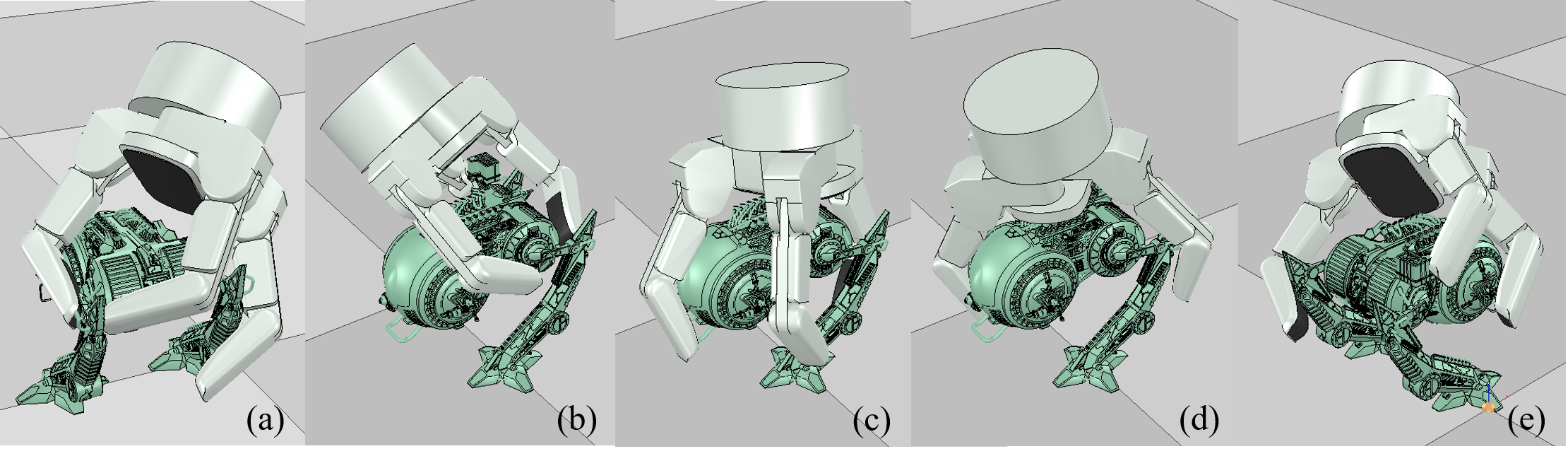}
		\caption{Visualization of 5 out of 7 grasps found on robot object (Arranged with quality from high to low). }
		\label{fig:sim_vis}
	\end{center}
\end{figure}

\begin{figure}[t]
	\begin{center}
		\includegraphics[width=3.3in]{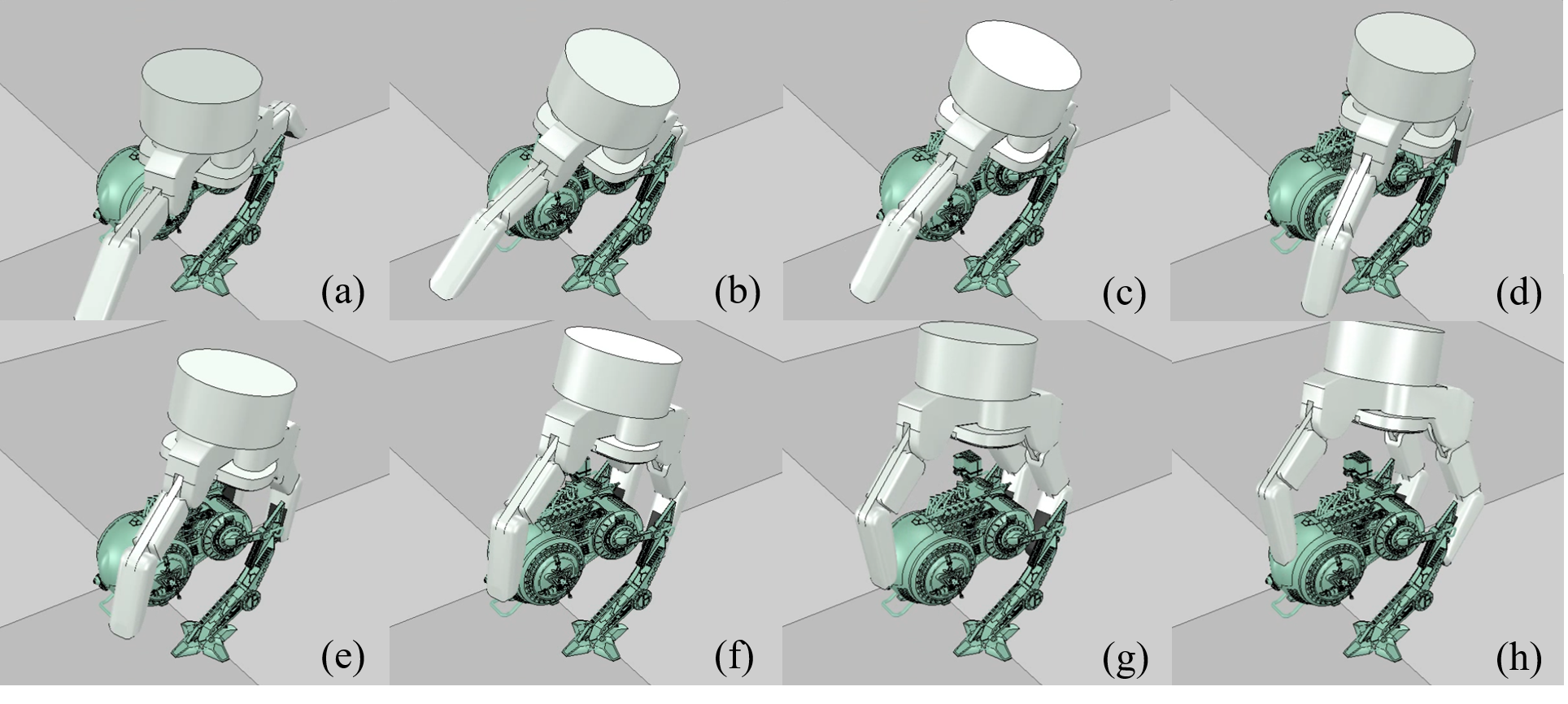}
		\caption{Simulation result of the iterative PPO-JPO on robot object. Snapshots from (a) to (h). }
		\label{fig:sim_ani}
	\end{center}
\end{figure}
Figure~\ref{fig:sim_vis} shows 5 out of 7 collision-free grasps found with 10 samples in simulation using the proposed PPO-JPO method. The object to be grasped was a \textit{robot} model with a complex shape. The proposed algorithm was able to find the versatile grasps without colliding with the object and the ground. The majority of the grasps found were precision grasps with fingertip contact. 

Figure~\ref{fig:sim_ani} shows the animation of a grasp search on the \textit{robot} object. The iterative PPO-JPO started searching from a vertical grasp with fully opened hand (Fig.~\ref{fig:sim_ani}(a)) and gradually adjusted the palm pose/joints angles to maximize the quality and avoid the collision, as shown in Fig.~\ref{fig:sim_ani}(b-h).  
\begin{figure}[t]
	\begin{center}
		\includegraphics[width=3.3in]{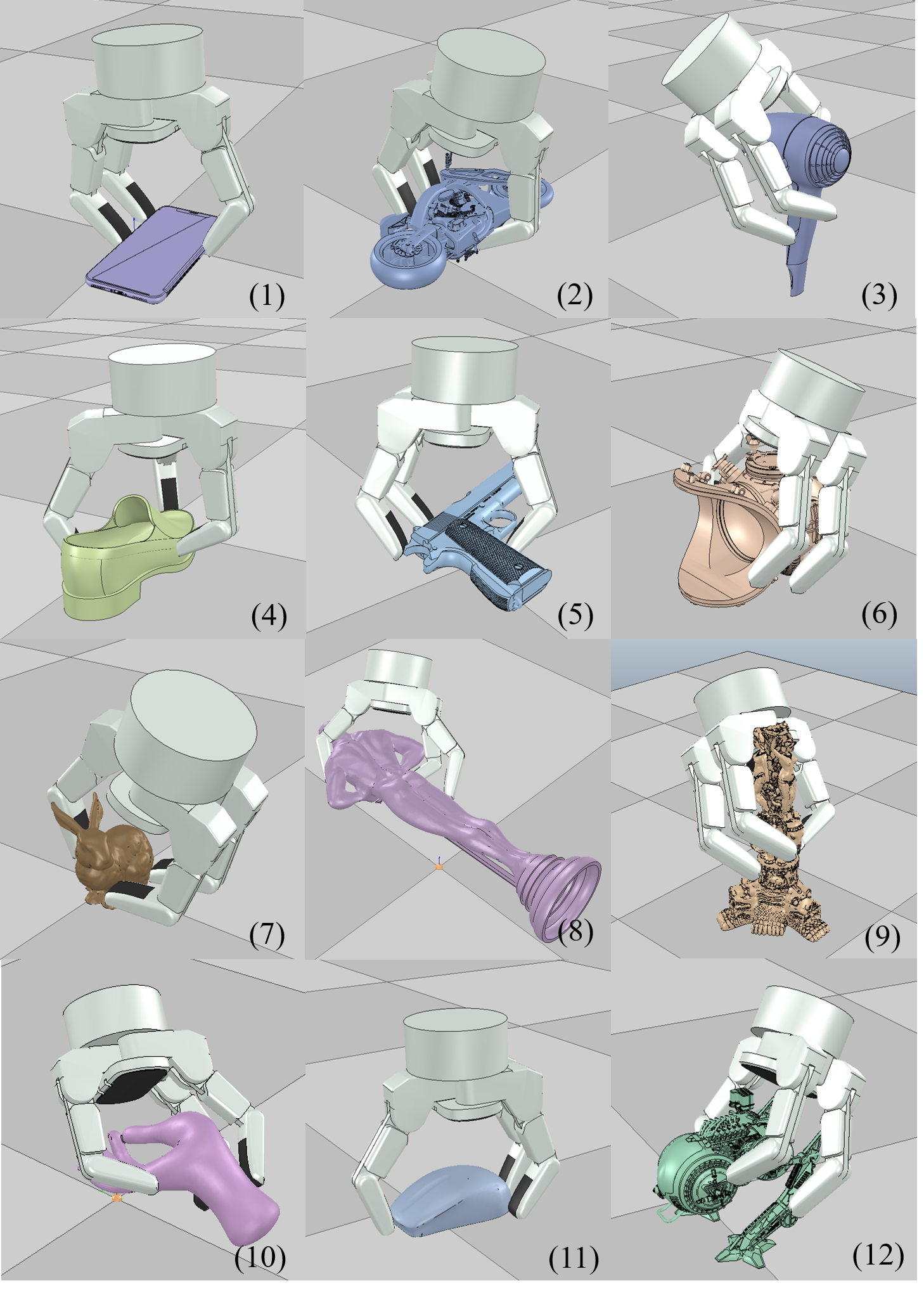}
		\caption{Visualization of the iterative PPO-JPO on 12 different objects. }
		\label{fig:sim_vis_12}
	\end{center}
\end{figure}

The grasp results on 12 different objects are shown in Figure~\ref{fig:sim_vis_12}. The iterative PPO-JPO was able to find collision-free precision grasps for a) thin objects close to the ground (Fig.~\ref{fig:sim_vis_12}(1,5)), b) objects with complex surfaces (Fig.~\ref{fig:sim_vis_12}(2,6,9,12)), or c) the objects with sharp edges (Fig.~\ref{fig:sim_vis_12}(3,7,12)).

The numerical results of the iterative PPO-JPO on these objects are shown in Table~\ref{tab:opt_details}.  Among the 12 different objects, the \textit{iPhone X}, \textit{gun}, \textit{hand} and \textit{mouse} are among the most challenging objects to grasp. The proposed algorithm was not sensitive to object complex surfaces but rather suffered from the low-height properties of the objects due to the collisions caused by the narrow space between the feasible grasp region and the ground. 

The qualities used in iterative PPO-JPO were a geometric metric. To reflect the physical conditions of the grasp, we also ranked the grasps with the physical grasp quality proposed in~\cite{fan2019efficient}. 
In average, the iterative PPO-JPO were able to find 6.58 collision-free grasps out of 10 samples within 3.26 secs (0.496 sec/grasp). 

\begin{table}[t]
		\centering
		\caption{Numerical Results of the iterative PPO-JPO}
		\label{tab:opt_details}
	\begin{tabular}{l|l|l|l|l}
		\hline
		{ID} & {Object} & {\begin{tabular}[c]{@{}c@{}} collision-free\#\\ \hline total samples \end{tabular} } & {Time (s)} &{Qualities~\cite{fan2019efficient}}       \\\hline\hline
		1& iPhone X&           5/10&         2.48&     -1.74   \\
		2& Motorbike&      6/10&     3.72&          1.01 \\
		3& Hairdryer&          10/10&      3.44&          12.52 \\
		4& Shoe&           6/10&      3.45&          1.20 \\ 
		5& Gun&              4/10&   2.89&        -4.20 \\
		6& Diving helmet&         9/10&      3.85&          10.27 \\
		7& Bunny&             8/10&    3.15&          7.51 \\
		8& Oscar&           8/10&   3.60&          5.11 \\
		9& Catcam&              7/10&   3.67&          2.98\\
		10& Hand&               5/10&    3.15&        -1.59\\
		11& Mouse&              4/10&    2.48&        -4.39 \\
		12& Robot&              7/10&    3.27&        4.70
		  \\\hline \hline
	1-12	& Average&          6.58/10&  3.263&      2.782      
	\end{tabular}
\end{table}

\begin{figure}[t]
	\begin{center}
		\includegraphics[width=3.6in]{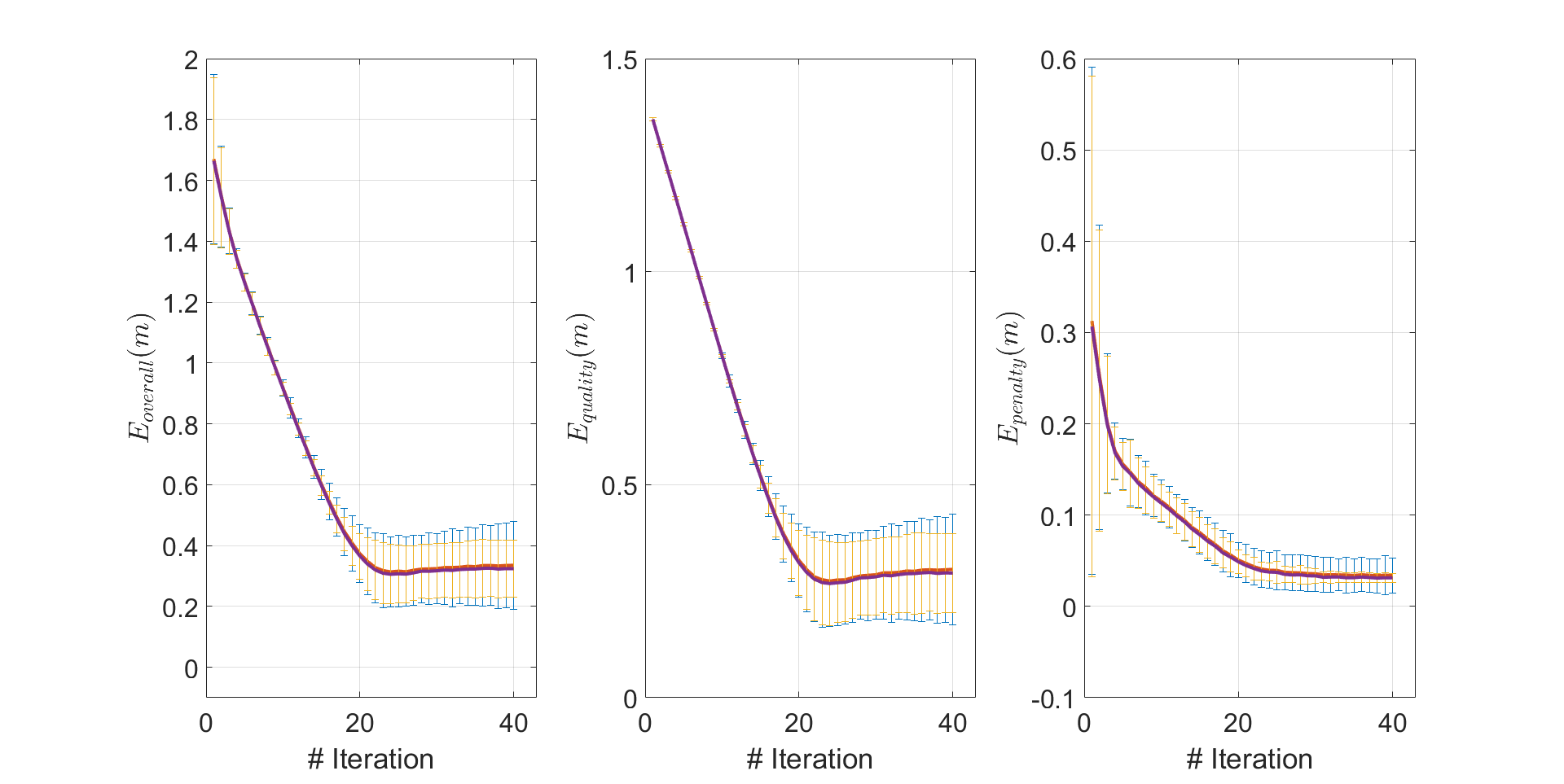}
		\caption{Error profile of iterative PPO-JPO on \textit{Bunny} object. (a) Overall error without multiplying penalty. (b) Quality error including $Q_{com}$, $Q_{jc}$ and $Q_{align}$. (c) Penalty error including $Q_{col}$ and $Q_{cls}$ with multiplying penalty. The red (purple) and blue (yellow) plots show the mean and standard deviation for all (collision-free) grasps. }
		\label{fig:sim_bunny_errors}
	\end{center}
\end{figure}

Figure~\ref{fig:sim_bunny_errors} shows the error reduction profile running the iterative PPO-JPO algirthm on bunny objects. The algorithm ran for $50$ trials, and generated $46$ collision-free grasps. We recorded the quality error $E_{quality} = -Q_{com} - Q_{jc} - Q_{alig}$ and the penalty error $E_{penalty} = E_{col} + E_{cls}$. In average, the grasp quality $E_{quality}$ reduced from $1.36\pm 0.0036$ m to $0.29 \pm 0.093$ m, and the penalty error $E_{penalty}$ reduced from $0.31 \pm 0.27$ m to $0.031 \pm 0.0050$ m.

\begin{figure*}[t]
	\begin{center}
		\includegraphics[width=5.55in]{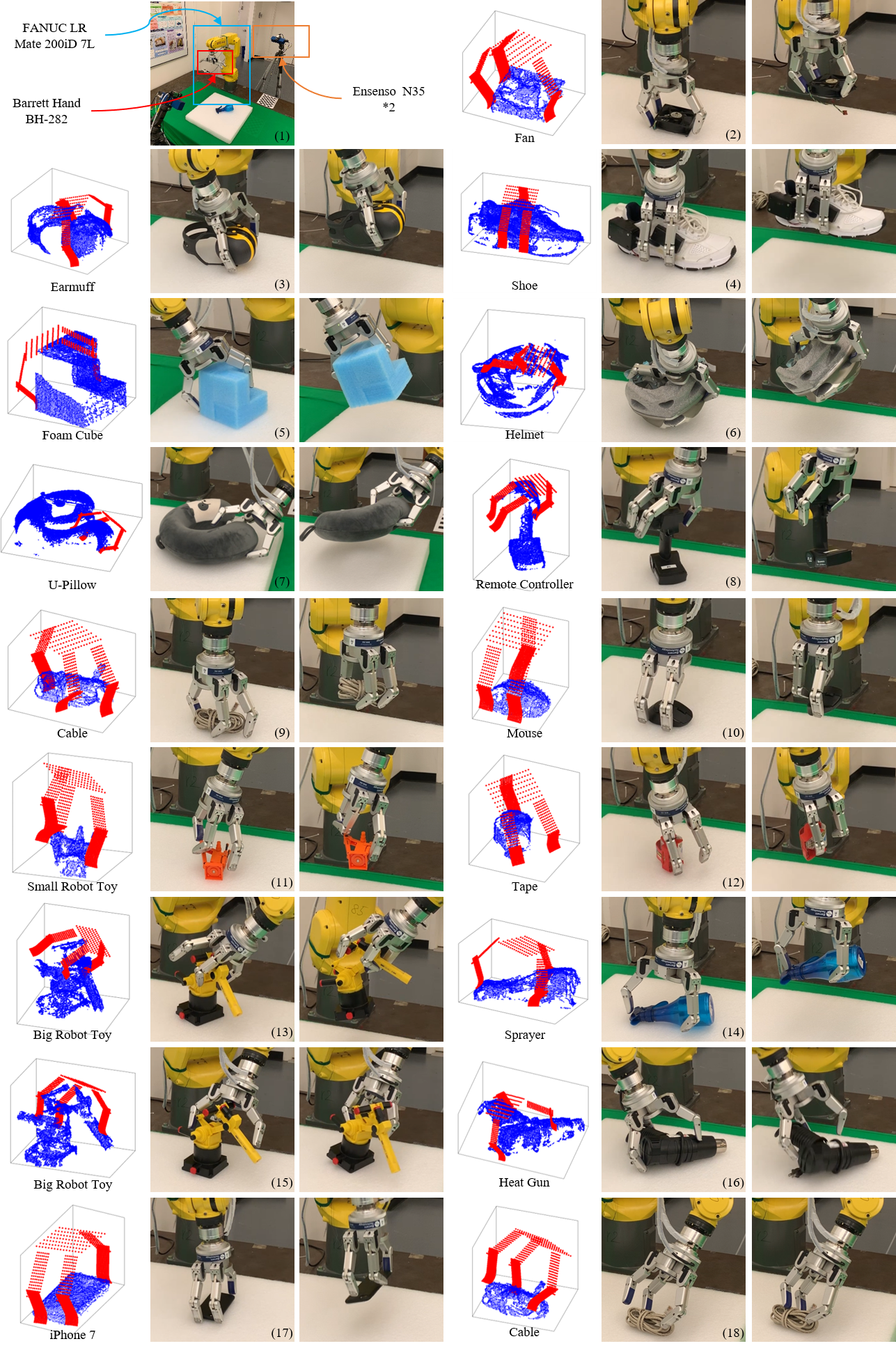} 
		\caption{(1) Experimental Setup and (2-18) The grasp planning and execution results on 15 objects. Subfigure (2-14) are successful grasp trials, and (15-18) are failure cases. For each subfigure, (Left) shows perceived point cloud and planned grasps, (Middle) shows physical grasps reaching the target grasp, and (Right) shows the execution results. }
		\label{fig:results_fig}
	\end{center}
\end{figure*}

\subsection{Experiment Results}
This section shows the experimental results with the BarrettHand BH8-282. To generate the trajectory to approach the target grasp, we used the grasp trajectory optimization (GTO) proposed in our previous work~\cite{fan2019efficient}.  Therefore, the whole experiment can be separated into three phases: 1) the grasp planning by iterative PPO-JPO, 2) the grasp trajectory generation by~\cite{fan2019efficient}, and 3) the object clamping by simply continuing finger motion for certain time until the tactile sensor reading reaching the target value ($1.0N/cm^2$).  The focus of this paper is grasp planning using the proposed iterative PPO-JPO algorithm. 

Figure~\ref{fig:results_fig} shows the experimental setup (Fig.~\ref{fig:results_fig}(1)), the grasp planning and execution results (Fig.~\ref{fig:results_fig}(2-18)) on 15 different objects. The perceived point cloud and the located grasp are shown on the left side of each subfigure. The physical grasp pose and the execution result of the planned grasp are shown in the middle and right, respectively. 

To address the noise and incompleteness of the perceived point cloud, the contact points were regarded as a region close to the fingertip, instead of a single point. Therefore, the system exhibits certain robustness to the noise and incompleteness of the object point cloud, as shown in Fig.~\ref{fig:results_fig}(2-14).  

While the algorithm shows certain robustness to noise and incompleteness of the point cloud, the resistance to uncertainties remains a challenge. These uncertainties were 1) positioning uncertainties including calibration error ($\sim 3$ mm for robot-camera frame alignment), installation error ($\sim 1^\circ$ TCP-palm alignment), actuation error ($\sim 2.0^\circ$ finger joint tracking error), 2) communication uncertainties including synchronization error ($\sim 0.1$ sec ROS-Matlab transmission misalignment), non-real time error ($\sim 0.1$ sec ROS latency on different fingers), and dynamics uncertainties including the mass uncertainties, friction uncertainties and softness uncertainties. 

Figure~\ref{fig:results_fig}(15-18) shows four failure cases caused by these uncertainties. More specifically, Figure~\ref{fig:results_fig}(15) was failed from the unsynchronized contacts of different fingers. The finger which contacted with the object first would keep pushing the object, introducing extra disturbance and perturbing the object. Consequently, the hand contacted with the object on undesired positions and caused the object slipped.  
Figure~\ref{fig:results_fig}(16) slipped off since the object was heavy and the plug was soft. However, the quality metric cannot take these factors into account. Similar failure appears in Fig.~\ref{fig:results_fig}(17,18).


\section{Conclusion and Discussion} 
\label{sec:conclusion}

This paper has proposed an efficient optimization model for precision grasp planning. To optimize the quality and avoid the collision, the planning problem has been formulated into an optimization with penalties and solved by iterating between the palm pose optimization (PPO) and joint position optimization (JPO). The iterative PPO-JPO algorithm was able to locate a collision-free grasp within 0.50 sec (based on 120 grasps on 12 objects in different categories). Experiments on a physical BarrettHand further demonstrated the effectiveness of the algorithm. The experimental videos are available at~\cite{website}. 

The current implementation has several limitations. First, the force allocation for the post-grasp phase is absent. The current close-and-clamp method may introduce unexpected collision and extra disturbance in the post-grasp phase. The force optimization, however, can be costly due to the requirements of certain dexterity and extra sensors of individual fingers. Developing a post-grasp strategy to resist the external disturbance for the hands with limited dexterity and low-priced sensors is our future work. 

Second, the precision grasps generated by iterative PPO-JPO are less robust to uncertainties compared with the power grasps generated by~\cite{fan2019efficient}. This is intuitive since the precision grasps rely on fingertip (frictional) force to grasp the object, and a small contact mismatch introduced by uncertainties will produce a large force error. On the other hand, the power grasps tend to enclose the object with the whole hand and generate form closure grasps with large contact regions, and the contact mismatch shifts contact regions without producing failure. The future work will introduce robustness index (i.e. flatness of contact surfaces) to search for more robust grasps.

\addtolength{\textheight}{-1cm}   



 

\bibliographystyle{IEEEtran}
\bibliography{references}

\begin{thebibliography}{10}
\providecommand{\url}[1]{#1}
\csname url@samestyle\endcsname
\providecommand{\newblock}{\relax}
\providecommand{\bibinfo}[2]{#2}
\providecommand{\BIBentrySTDinterwordspacing}{\spaceskip=0pt\relax}
\providecommand{\BIBentryALTinterwordstretchfactor}{4}
\providecommand{\BIBentryALTinterwordspacing}{\spaceskip=\fontdimen2\font plus
\BIBentryALTinterwordstretchfactor\fontdimen3\font minus
  \fontdimen4\font\relax}
\providecommand{\BIBforeignlanguage}[2]{{%
\expandafter\ifx\csname l@#1\endcsname\relax
\typeout{** WARNING: IEEEtran.bst: No hyphenation pattern has been}%
\typeout{** loaded for the language `#1'. Using the pattern for}%
\typeout{** the default language instead.}%
\else
\language=\csname l@#1\endcsname
\fi
#2}}
\providecommand{\BIBdecl}{\relax}
\BIBdecl

\bibitem{website}
{Experimental Videos for Planning Precision Grasps with Multi-Fingered Hands},
  {http://me.berkeley.edu/\%7Eyongxiangfan/IROS2019/ppojpo.html}.

\bibitem{ferrari1992planning}
C.~Ferrari and J.~Canny, ``Planning optimal grasps,'' in \emph{Robotics and
  Automation, 1992. Proceedings., 1992 IEEE International Conference on}.\hskip
  1em plus 0.5em minus 0.4em\relax IEEE, 1992, pp. 2290--2295.

\bibitem{hang2016hierarchical}
K.~Hang, M.~Li, J.~A. Stork, Y.~Bekiroglu, F.~T. Pokorny, A.~Billard, and
  D.~Kragic, ``Hierarchical fingertip space: A unified framework for grasp
  planning and in-hand grasp adaptation,'' \emph{IEEE Transactions on
  robotics}, vol.~32, no.~4, pp. 960--972, 2016.

\bibitem{aleotti2011part}
J.~Aleotti and S.~Caselli, ``Part-based robot grasp planning from human
  demonstration,'' in \emph{2011 IEEE International Conference on Robotics and
  Automation}.\hskip 1em plus 0.5em minus 0.4em\relax IEEE, 2011, pp.
  4554--4560.

\bibitem{vahrenkamp2018planning}
N.~Vahrenkamp, E.~Koch, M.~W{\"a}chter, and T.~Asfour, ``Planning high-quality
  grasps using mean curvature object skeletons,'' \emph{IEEE Robotics and
  Automation Letters}, vol.~3, no.~2, pp. 911--918, 2018.

\bibitem{dang2012semantic}
H.~Dang and P.~K. Allen, ``Semantic grasping: Planning robotic grasps
  functionally suitable for an object manipulation task,'' in \emph{2012
  IEEE/RSJ International Conference on Intelligent Robots and Systems}.\hskip
  1em plus 0.5em minus 0.4em\relax IEEE, 2012, pp. 1311--1317.

\bibitem{huang2013learning}
B.~Huang, S.~El-Khoury, M.~Li, J.~J. Bryson, and A.~Billard, ``Learning a real
  time grasping strategy,'' in \emph{Robotics and Automation (ICRA), 2013 IEEE
  International Conference on}.\hskip 1em plus 0.5em minus 0.4em\relax IEEE,
  2013, pp. 593--600.

\bibitem{osa2016experiments}
T.~Osa, J.~Peters, and G.~Neumann, ``Experiments with hierarchical
  reinforcement learning of multiple grasping policies,'' in
  \emph{International Symposium on Experimental Robotics}.\hskip 1em plus 0.5em
  minus 0.4em\relax Springer, 2016, pp. 160--172.

\bibitem{varley2015generating}
J.~Varley, J.~Weisz, J.~Weiss, and P.~Allen, ``Generating multi-fingered
  robotic grasps via deep learning,'' in \emph{Intelligent Robots and Systems
  (IROS), 2015 IEEE/RSJ International Conference on}.\hskip 1em plus 0.5em
  minus 0.4em\relax IEEE, 2015, pp. 4415--4420.

\bibitem{ciocarlie2007dexterous}
M.~Ciocarlie, C.~Goldfeder, and P.~Allen, ``Dexterous grasping via eigengrasps:
  A low-dimensional approach to a high-complexity problem,'' in \emph{Robotics:
  Science and Systems Manipulation Workshop-Sensing and Adapting to the Real
  World}.\hskip 1em plus 0.5em minus 0.4em\relax Citeseer, 2007.

\bibitem{fan2018real}
Y.~Fan, T.~Tang, H.-C. Lin, and M.~Tomizuka, ``Real-time grasp planning for
  multi-fingered hands by finger splitting,'' in \emph{2018 IEEE/RSJ
  International Conference on Intelligent Robots and Systems (IROS)}.\hskip 1em
  plus 0.5em minus 0.4em\relax IEEE, 2018, pp. 4045--4052.

\bibitem{fan2018grasp}
Y.~Fan, H.-C. Lin, T.~Tang, and M.~Tomizuka, ``Grasp planning for customized
  grippers by iterative surface fitting,'' in \emph{2018 IEEE 14th
  International Conference on Automation Science and Engineering (CASE)}.\hskip
  1em plus 0.5em minus 0.4em\relax IEEE, 2018, pp. 28--34.

\bibitem{schulman2013finding}
J.~Schulman, J.~Ho, A.~X. Lee, I.~Awwal, H.~Bradlow, and P.~Abbeel, ``Finding
  locally optimal, collision-free trajectories with sequential convex
  optimization.'' in \emph{Robotics: science and systems}, vol.~9, no.~1.\hskip
  1em plus 0.5em minus 0.4em\relax Citeseer, 2013, pp. 1--10.

\bibitem{fan2019efficient}
Y.~Fan and M.~Tomizuka, ``Efficient grasp planning and execution with
  multi-fingered hands by surface fitting,'' \emph{arXiv preprint
  arXiv:1902.10841}, 2019.

\end{thebibliography}

\end{document}